# Easy Data Augmentation in Sentiment Analysis of Cyberbullying

Alwan Wirawan
Department of Informatics
Faculty of Information Technology and Science Data
Universitas Sebelas Maret
Surakarta, Indonesia
alwanwrwn@student.uns.ac.id

Hasan Dwi Cahyono
Department of Informatics
Faculty of Information Technology and Science Data
Universitas Sebelas Maret
Surakarta, Indonesia
hasandc@staff.uns.ac.id

Winarno
Department of Informatics
Faculty of Information Technology and Science Data
Universitas Sebelas Maret
Surakarta, Indonesia
win@staff.uns.ac.id

*Abstract*—Instagram, a social media platform, has in the vicinity of 2 billion active users in 2023. The platform allows users to post photos and videos with one another. However, cyberbullying remains a significant problem for about 50% of young Indonesians. To address this issue, sentiment analysis for comment filtering uses a Support Vector Machine (SVM) and Easy Data Augmentation (EDA). EDA will augment the dataset, enabling robust prediction and analysis of cyberbullying by introducing more variation. Based on the tests, SVM combination with EDA results in a 2.52% increase in the k-Fold Cross Validation score. Our proposed approach shows an improved accuracy of 92.5%, 2.5% higher than that of the existing state-of-the-art method. To maintain the reproducibility and replicability of this research, the source code can be accessed at uns.id/eda_svm.

## I. Introduction

Instagram is a popular social media site enabling users to share photographs and videos. There are approximately 2 billion active Instagram users by 2023. The active users consist of various groups of ages, from children to adults. Further, the purpose of these users can vary, namely displaying personal profiles for ordinary people and artists and business promotion tools [1]. Regardless of the intention, the majority of Instagram users are still unaware that online interactions can be harmful and can cause cyberbullying – bullying that occurs on the internet. Even interactions in the form of ratings or comments can also be considered cyberbullying without the users of Instagram being aware of it by Instagram users. This lack of awareness warns users, parents, relatives, and governments to reduce this harmful behavior.

According to UNICEF, at least 50% of teenagers in Indonesia have experienced cyberbullying. In fact, nearly 22.4% of the 161 violent cases reported to the Indonesian Child Protection Commission (KPAI) in 2018 involved cyberbullying [2]. To mitigate the impact of cyberbullying in social media, sentiment analysis is utilized to filter comments. Sentiment analysis will then determine which comments are good and which are not. A non-offensive comment contains positive words and has a positive purpose, while vice versa for an offensive comment. Comment filtering can be an effort to prevent cyberbullying victims from being exposed to harmful comments after being filtered. Thus, the emotional and psychological damage caused by cyberbullying can be minimized [3].

A previous study has explored sentiment analysis to prevent cyberbullying and used the Support Vector Machine (SVM) [4]. Using 400 heterogeneous cases as the dataset, the SVM algorithm is used to carry out a sentiment analysis on comment objects on Instagram and reported to have a promising result with an accuracy of 90%. The performance of this model is promising and can potentially be improved, especially when the dataset is small (less than 1280 samples per class). With a small dataset for training, a model is more likely to learn the specifics of the training data than that of the underlying patterns, as the model has not seen enough examples of different ways to represent data and maintains the model generality on the testing dataset. Thus, understanding underlying patterns is essential when classifying new data [5].

To understand the patterns, two studies compare augmentation methods across various datasets. The first study [6] exercises a generative model that combines Variational Autoencoding (VAE) with prior and posterior sampling to produce fake data. The result of this study is a 0.04% and 2.02% increase in two different data sets. The second approach, employing a two-way language model to replace words with other words inferred from the context of sentences, led to an increase of 0.5% in five datasets [7]. To train a VAE or a two-way language model, however, the computational costs are high, which limits the implementation of these models. Easy Data Augmentation (EDA), which requires no model language training and external data sets, produces comparable results to the VAE and two-way language models while maintaining its computational simplicity [8] by creating new synthetic data. Synthetic data is instrumental in Natural Language Processing in which data can be scarce or difficult to obtain, while models must be able to handle multiple inputs and variations.

Therefore, in this research, we explore using SVM as a machine learning [4] and investigate the EDA method to augment the data. EDA will create a new sample by transforming the existing ones. To select the optimal results, sample transformations are also investigated, in particular by adding synonyms, swapping two words in one sentence, or deleting one of the words. By applying these techniques to the existing data, we can create new synthetic data that can be utilized to train models more effectively.

## II. Related Work

This research follows [4] for the classification of comment sentiment framework related to cyberbullying on Instagram using SVM. With training data composed of 50% test and 50% training data, this study reflects some promising results with 90% accuracy, 94.44% precision, 85% recall, and 89.47% f-measure, respectively. More recent research by [9] investigated that the wording of social media comments is inconsistent and noisy. The study uses SVM as the classification algorithm and evaluates the performance with the best accuracy results of 81.06%.

In different studies about aspect-based sentiment analysis, the limitation of the dataset has been improved by introducing EDA. EDA increased the accuracy of the two test data sets by 1 - 0.5 %, respectively [10]. However, the results were reported using the SemEval 2016 dataset with English sentences. Thus, the performance has not yet been explored for cyberbullying in Bahasa Indonesia.

III. METHODOLOGY

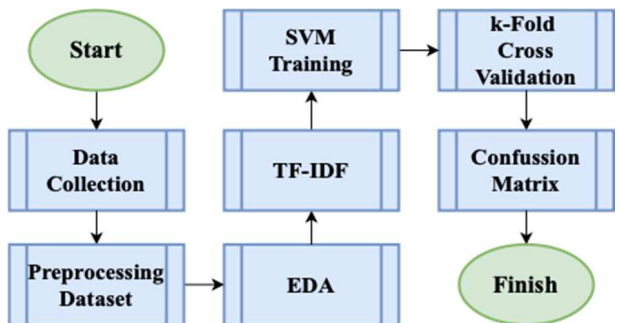

Fig. 1. The flow of our study.

A. Data Collection

The dataset used in Fig. 1 is the same as [4]. The data is downloaded from here https://github.com/rizalespe/Dataset-Sentimen-Analisis-Bahasa-indonesia/. The dataset consists of 400 records with two classes, positive or negative. Here, negative means cyberbullying, and positive means not cyberbullying. In total, each class has 200 samples equally.

B. Preprocessing Dataset

Preprocessing is the first step in text mining, and it involves converting the data into the desired format. This procedure is used to investigate, process, and organize data. Further, we can examine textual relationships between structured and unstructured data using preprocessing [11].

- Case Folding and Data Cleaning

    Case Folding is the first step during preprocessing and attempts to standardize word forms in lowercase [12]. The text cleaning technique is the subsequence step that removes delimiting commas (,), periods (.), and other punctuation marks. Data cleaning aims to decrease noise in the data [13].

- Language Normalization

    Language normalization is then performed on non-standard words at the preprocessing step. According to the "Kamus Besar Bahasa Indonesia" (KBBI). This stage attempts to restore the written form of each word. This procedure matches each word in the training data document and test data to a term in a non-standard language dictionary in this procedure. Language normalization is done with Spacy's lemmatization [14].

- Stopword Removal

    Stopwords are a collection of frequently used words that have no meaning and are rarely used. Common words will be omitted as part of this procedure to limit the number of words stored by the system [12]. The stop word list was taken from Spacy's with 757 words.

- Stemming

    Stemming identifies the stem (base words) of words removed by stopwords (filtering). The dictionary method and the rule approach are two approaches to stemming. The stemming process uses a stemmer from sastrawi[15].

- Tokenization

    Tokenization is a method of converting text into tokens that function as basic units. Whitespace will be deleted because of this procedure. By breaking text into tokens, tokenization facilitates further analysis, such as text classification and sentiment analysis [12].

C. Easy Data Augmentation (EDA)

At the EDA stage, a data augmentation process is carried out for SVM training. EDA is taken from https://github.com/jasonwei20/eda_nlp. For Synonym Replacement and Random Insertion operations, we facilitate the use of thesaurus from https://github.com/victoriasovereigne/tesaurus. This algorithm will randomly choose one of the following operations and apply it to a specific sentence from the training data. EDA has four distinct operations, namely:

1. Synonym Replacement (SR) - Choose n words randomly from the sentence. Each of these words must be replaced with its synonym at random.

2. Random Insertion (RI) - Find random synonyms for random words that are not stopwords in the sentence. Place the synonyms in random parts of the text. Repeat this process n times.

3. Random Swap (RS) - In a sentence, swap the positions of two random words. Do this n times.

4. Random Deletion (RD) - In the sentence, randomly delete each word using probability p.

The EDA has five parameters for each α and $n_{aug}$ for the desired sample augmentation data [8]. Consequently, we will investigate 25 different scenarios.,

D. TF-IDF

A numerical statistic called TF-IDF is used to determine the importance of a term in a document through text mining and information retrieval. Important keywords are extracted, document rankings are carried out, and text-based model functionality is improved [16]. To support the reliability and reproducibility of this research, we utilize modules from sci-kit-learn for TF-IDF.

Term Frequency (TF) quantifies the occurrence of terms in documents. About the number of words in the document, TF displays the frequency of terms. On the other hand, Inverse Document Frequency (IDF) measures the inverse of the frequency of terms throughout document collections Eq. (1) – (2).

$$W_{tf_{t,d}} = \begin{cases} 1 + log_{10} \, tf_{t,d}, & if \, tf_{t,d} > 0 \\ 0 \end{cases} \quad (1)$$

$$idf_t = log_{10} \frac{N}{df_t} \quad (2)$$

Multiplying the term's TF value by the IDF score will produce the TF-IDF score. The higher the TF-IDF score of a term in a document, the greater its significance and relevance to the document, and increases the precision of text-based models and systems.

*E. SVM Training*

The utilization of SVM in this research is intended to do sentiment analysis. SVM is a problem-solving algorithm that can solve linear and non-linear problems. The kernel solves non-linear problems in high-dimensional workspaces by finding the hyperplane that maximizes the margin between data classes. The hyperplane is essential for distinguishing between two groups, class +1 and class -1, each of which has a unique pattern [4]. SVM uses the classifier from the widely accessible and popular sci-kit-learn modules. A grid search will be used to find the best combination of parameters. Grid search will try all possibilities from the desired parameter range.

The K kernel function $K(x_i, x_j)$ is used to make decisions in the SVM method, with $x_i, x_j$ as a pair of data from training data and degree d to control the complexity and flexibility of the decision boundary [4]. The polynomial kernel will be used in this study, as shown in Eq. (3):

$$K(x_i, x_j) = (x_i \cdot x_j + C)^d \quad (3)$$

*F. Evaluation using k-Fold Cross Validation*

k-fold cross-validation is a resampling technique for assessing model performance on data, where k is the number of folds. First, the model is trained on section k-1 and evaluated on the remaining parts. The data is then divided into k sections. Next, each piece was used as test data during this process, repeated k number of times. Finally, the average k score is used as the final estimate of the model performance [9]. Experiments will be carried out until the best model is found.

The number of k is the amount of fold that can be set to measure model performance. A larger k will provide a more accurate estimate of model performance and be more computationally expensive. So, deciding a suitable value of k is of interest.

*G. Confusion Matrix*

The confusion matrix is a performance assessment technique for evaluating a classification model's accuracy. By contrasting the projected labels with the actual labels, the assessment summarizes the classification issue predictions made by the model [17].

IV. RESULT AND DISCUSSION

*A. Research Environment*

The research was conducted using a computer device. Computer hardware specifications, namely Apple M1 with 8 CPU cores, 8 GPU cores, 16 neural engines, and 8 GB of RAM. The operating system in computer devices uses macOS Sonoma.

*B. Easy Data Augmentation*

- Synonym Replacement (SR)

*Table 1. Synonym replacement result*

| input | jujur aja ya ni org bagus kaga oplas tu foto dia sblm dn oplas dioplas muka **aneh** kyknya gagal oplasnya |
|---|---|
| replaced word | 'aneh' with 'heran' |
| output | juju raja ya ni org bagus kaga oplas tu foto sblm dn oplas dioplas muka **heran** kyknya gagal oplasnya |

Based on Table 1 and the author's observation of the resulting data, the results of data augmentation using SR are the best among the others because SR replaces words with their synonyms and does not change the sentence structure. The meaning of the sentence remains consistent. Subsequently, the labels stay true while introducing accurate new features.

- Random Insertion (RI)

*Table 2. Random insertion result*

| input | jujur aja ya ni org bagus kaga oplas tu **foto** dia sblm dn oplas dioplas muka aneh kyknya gagal oplasnya |
|---|---|
| inserted word | 'cetakan' ('foto's synonym) |
| output | juju raja ya ni org **cetakan** bagus kaga oplas tu foto sblm dn oplas dioplas muka aneh kyknya gagal oplasnya |

An example of data augmentation results using RI is shown in Table 2. Based on the author's observation of the results, this operation is suitable for adding data variations by adding a new word. Even though the random placement makes the word seem out of place in the context of the sentence.

- Random Swap (RS)

*Table 3. Random swap result*

| input | **jujur** aja ya ni org bagus kaga **oplas** tu foto dia sblm dn oplas dioplas muka aneh kyknya gagal oplasnya |
|---|---|
| swapped word | 'jujur' with 'oplas' |
| output | **oplas** aja ya ni org bagus kaga **jujur** tu foto sblm dn oplas dioplas muka aneh kyknya gagal oplasnya |

Based on Table 3, adding data with RS operation can alter the meaning of the preceding statement. From the sentence above, the photo implies the opposite even when an "oplas" was performed with good results. This meaning differs from the input sentence, which means the photo in question provides evidence that "oplas" failed and the results were disappointing.

- Random Deletion (RD)

*Table 4. Random deletion result*

| input | jujur aja ya ni org bagus kaga oplas tu foto dia sblm dn oplas dioplas muka aneh kyknya gagal oplasnya |
|---|---|
| deleted word | 'dia' |
| output | oplas aja ya ni org bagus kaga jujur tu foto sblm dn oplas dioplas muka aneh kyknya gagal oplasnya |

Table 5. TF-IDF results

| term(t) | tf | $w_{tf}$ | df | idf | tf-idf |
|---|---|---|---|---|---|
| jujur | 1 | 1 | 6 | 1.78 | 1.78 |
| aja | 1 | 1 | 58 | 0.79 | 0.79 |
| ya | 1 | 1 | 66 | 0.74 | 0.74 |
| ni | 1 | 1 | 13 | 1.44 | 1.44 |
| org | 1 | 1 | 35 | 1.01 | 1.01 |
| bagus | 1 | 1 | 17 | 1.33 | 1.33 |
| kaga | 1 | 1 | 3 | 2.08 | 2.08 |
| oplas | 2 | 1.301 | 5 | 1.86 | 2.42 |
| tu | 1 | 1 | 10 | 1.56 | 1.56 |
| foto | 1 | 1 | 5 | 1.86 | 1.86 |
| dia | 1 | 1 | 3 | 2.08 | 2.08 |
| sblm | 1 | 1 | 2 | 2.26 | 2.26 |
| dn | 1 | 1 | 2 | 2.26 | 2.26 |
| dioplas | 1 | 1 | 3 | 2.08 | 2.08 |
| muka | 1 | 1 | 18 | 1.30 | 1.30 |
| aneh | 1 | 1 | 3 | 2.08 | 2.08 |
| kyknya | 1 | 1 | 3 | 2.08 | 2.08 |
| gagal | 1 | 1 | 4 | 1.95 | 1.95 |
| oplasnya | 1 | 1 | 3 | 2.08 | 2.08 |

Table 6. The parameters of SVM

| C | γ | Kernel | Degree | Random State |
|---|---|---|---|---|
| 0.0001 | 100 | Polynomial | 2 | 42 |

The outcomes of data augmentation using RD operation are satisfactory, according to Table IV and the author's observations of the generated data, albeit occasionally. Inconsequential words significantly impact a sentence, as in the input sentence when the word "dia" is eliminated. The pronoun "dia" in the input sentence emphasizes that the item in the photo belongs to the subject, a person who has undergone plastic surgery.

### C. TF-IDF

The word–weighting process is carried out separately for training and testing data. This separation is intended so that the resulting model passes over the features in the testing data. As a result, the model provides optimistic performance. Table 5 shows the results of TF-IDF calculations for sample data.

TABLE I.

### D. SVM Training and k-Fold Cross Validation

The parameters used in the training process are obtained from the grid search results. The training results are evaluated using k=10. The optimal value of each parameter can be seen in Table 6 in line with the proposed initial [4].

TABLE II.

The training results using the ideal SVM parameters can be seen in Table 7. The value obtained will be the baseline (SVM only) for comparison after the data augmentation. Each model is compared using the accuracy matrix and 10-fold value.

Data augmentation is only done on training data. The baseline comprises training and testing data of 90:10 proportion. Hence, the baseline augments 360 data. This study uses the parameters $n_{aug}$ = {1, 2, 4, 8, 16, 32} and α = {0.05, 0.1, 0.2, 0.3, 0.4, 0.5} [8]. Therefore, there are 36 scenarios resulting from two parameter combinations. The results of each scenario can be seen in Table 8.

From the experiments conducted, the highest 10-fold value is shown in the combination of $n_{aug}$ = 16 and α = 0.1 of 89.74%, an increase of 2.52% from the baseline. The highest accuracy was shown in various combinations, amounting to 92.5%, an increase of 2.5% from the baseline.

TABLE III.

TABLE IV.

### E. Analysis and Results Comparison with Previous Research

The research results using Easy Data Augmentation succeeded in surpassing and improving the performance of the previous method. A comparison of our investigation with the previous approach is presented in Table 9.

The model performance increases after data augmentation using EDA. The precision value decreased by 1.9% from the four metrics used, while other metrics have increased. The recall has the highest value of 7.5%. There is a 2.5% increase in accuracy from the previous study. The generalization of the model is also improved, as seen from the rise in the k-Fold value. The best performance was obtained at $n_{aug}$ = 16 and α = 0.1. Some different values of k for k-Fold cross-validation have been investigated, while the best results were achieved at k=10.

Table 7. The baseline without augmentation results

| Model | Accuracy | Precision | Recall | F1 Score | 10-fold |
|---|---|---|---|---|---|
| SVM only | 90% | 90% | 90% | 90% | 87.2% |

Table 8. The baseline with augmentation results

| $n_{aug}$ | α | 10-fold | accuracy |
|---|---|---|---|
| 1 | 0.05 | 87.01% | 92.5% |
|  | 0.1 | 88.83% | 90% |
|  | 0.2 | 89.53% | 90% |
|  | 0.3 | 87.48% | 92.5% |
|  | 0.4 | 87.24% | 90% |
|  | 0.5 | 86.54% | 85.00% |
| 2 | 0.05 | 88.49% | 92.5% |
|  | 0.1 | 88.69% | 90% |
|  | 0.2 | 87.85% | 92.5% |
|  | 0.3 | 86.85% | 92.5% |
|  | 0.4 | 84.00% | 87.50% |
|  | 0.5 | 85.87% | 90% |
| 4 | 0.05 | 87.39% | 92.5% |
|  | 0.1 | 87.52% | 92.5% |
|  | 0.2 | 87.11% | 92.5% |
|  | 0.3 | 87.26% | 90% |
|  | 0.4 | 87.17% | 92.5% |
|  | 0.5 | 84.33% | 87.50% |
| 8 | 0.05 | 88.78% | 92.5% |
|  | 0.1 | 88.39% | 92.5% |
|  | 0.2 | 87.82% | 92.5% |
|  | 0.3 | 86.50% | 87.50% |
|  | 0.4 | 85.14% | 92.5% |
|  | 0.5 | 85.75% | 87.50% |
| 16 | 0.05 | 88.34% | 92.5% |
|  | 0.1 | **89.74%** | **92.5%** |
|  | 0.2 | 87.91% | 92.5% |
|  | 0.3 | 86.82% | 92.5% |
|  | 0.4 | 86.46% | 85.00% |
|  | 0.5 | 85.78% | 90% |
| 32 | 0.05 | 87.96% | 92.5% |
|  | 0.1 | 87.57% | 92.5% |
|  | 0.2 | 87.32% | 90% |
|  | 0.3 | 85.85% | 87.50% |
|  | 0.4 | 85.63% | 90% |
|  | 0.5 | 85.74% | 92.5% |

Table 9. The baseline comparison

| Model | Accuracy | Precision | Recall | F1 Score | 10-fold |
|---|---|---|---|---|---|
| Original[4] | 90% | 94.44% | 85% | 89.47% | - |
| SVM only | 90% | 90% | 90% | 90% | 87.22% |
| SVM+EDA | **92.5%** | 92.5% | **92.5%** | 92.5% | **89.74%** |

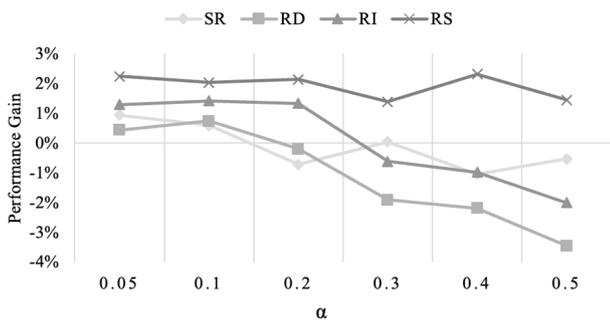

Fig. 2. The comparison of results for each EDA operation.

The performance increases from the four EDA processes (Fig. 2). For SR, improvement can be seen for small α values, but performance decreases for large α, most likely because changing too many words in a text varies the meaning. The performance gains for RI were more consistent across the range of values, possibly due to leaving out the sentences' original words and their relative order. RS results in significant performance improvements over all ranges of values because simply changing the word order results in similar meanings with different means of representation. RD had the most negligible performance gain compared to the others because word deletion can remove meaningful words from sentences.

V. CONCLUSION AND FUTURE WORKS

EDA research on cyberbullying sentiment analysis using the SVM method has been successfully carried out. The use of the EDA method improved the previous model's performance. The highest average k-Fold accuracy was obtained with $n_{aug}$ = 16 and α = 0.1 for the EDA parameter of 89.74%, which increased by 2.52% from baseline. The highest accuracy testing was obtained with $n_{aug}$ = 16 and α = 0.1 for the EDA parameter of 92.5%. Our approach goes over and above the previous research [4] with an increase in accuracy of 2.5%. This research can be considered as a reference to select the optimal parameters of EDA in Bahasa Indonesia.

Lemmatization during language normalization could be improved for more optimal results [18]. Some words are left unnormalized, causing data augmentation operations hardly optimal because of the use of synonyms. Unnormalized word synonym's could be beneficial to introduce new features in the model.

ACKNOWLEDGMENT

Many appreciation for the assistance provided by the Faculty of Information Technology and Science Data at Universitas Sebelas Maret.

REFERENCES

[1] J. A. Roberts and M. E. David, "Instagram and TikTok Flow States and Their Association with Psychological Well-Being," *Cyberpsychology, Behavior, and Social Networking*, vol. 26, no. 2, pp. 80–89, Feb. 2023, doi: 10.1089/cyber.2022.0117.

[2] Y. E. Riany and F. Utami, "Cyberbullying Perpetration among Adolescents in Indonesia: The Role of Fathering and Peer Attachment," *Int Journal of Bullying Prevention*, May 2023, doi: 10.1007/s42380-023-00165-x.

[3] J. Mula-Falcón and C. Cruz González, "Effectiveness of cyberbullying prevention programs on perpetration levels: a meta-analysis.," *revistafuentes*, pp. 12–25, 2023, doi: 10.12795/revistafuentes.2023.21525.

[4] W. A. Luqyana, I. Cholissodin, and R. S. Perdana, "Analisis Sentimen Cyberbullying pada Komentar Instagram dengan Metode Klasifikasi Support Vector Machine," vol. 2, pp. 4704–4713, Nov. 2018.

[5] L. Brigato and L. Iocchi, "A Close Look at Deep Learning with Small Data." arXiv, Oct. 25, 2020. Accessed: Jun. 19, 2023. [Online]. Available: http://arxiv.org/abs/2003.12843

[6] S. Qiu et al., "EasyAug: An Automatic Textual Data Augmentation Platform for Classification Tasks," in *Companion Proceedings of the Web Conference 2020*, Taipei Taiwan: ACM, Apr. 2020, pp. 249–252. doi: 10.1145/3366424.3383552.

[7] S. Kobayashi, "Contextual Augmentation: Data Augmentation by Words with Paradigmatic Relations." arXiv, May 16, 2018. Accessed: Jul. 27, 2023. [Online]. Available: http://arxiv.org/abs/1805.06201

[8] J. Wei and K. Zou, "EDA: Easy Data Augmentation Techniques for Boosting Performance on Text Classification Tasks," *arXiv:1901.11196 [cs]*, Aug. 2019, Accessed: Feb. 08, 2022. [Online]. Available: http://arxiv.org/abs/1901.11196

[9] A. W. Pradana and M. Hayaty, "The Effect of Stemming and Removal of Stopwords on the Accuracy of Sentiment Analysis on Indonesian-language Texts," *KINETIK*, pp. 375–380, Oct. 2019, doi: 10.22219/kinetik.v4i4.912.

[10] T. Liesting, F. Frasincar, and M. M. Trusca, "Data Augmentation in a Hybrid Approach for Aspect-Based Sentiment Analysis." arXiv, Mar. 29, 2021. Accessed: Jun. 12, 2023. [Online]. Available: http://arxiv.org/abs/2103.15912

[11] L. H. Lin, S. B. Miles, and N. A. Smith, "Natural Language Processing for Analyzing Disaster Recovery Trends Expressed in Large Text Corpora," in *2018 IEEE Global Humanitarian Technology Conference (GHTC)*, San Jose, CA: IEEE, Oct. 2018, pp. 1–8. doi: 10.1109/GHTC.2018.8601884.

[12] C. D. Manning, P. Raghavan, and H. Schütze, *Introduction to information retrieval*. New York: Cambridge University Press, 2008.

[13] A. Tabassum and R. R. Patil, "A Survey on Text Preprocessing & Feature Extraction Techniques in Natural Language Processing," *International Research Journal of Engineering and Technology (IRJET)*, vol. 7, no. 6, pp. 4864–4867, Jun. 2020.

[14] I. M. B. S. Darma, R. S. Perdana, and Indriati, "Penerapan Sentimen Analisis Acara Televisi Pada Twitter Menggunakan Support Vector Machine dan Algoritma Genetika sebagai MetodeSeleksi Fitur," vol. 2, no. 3, pp. 998–1007, Jan. 2018.

[15] M. Utomo, "Implementasi Stemmer Tala pada Aplikasi Berbasis Web," vol. 18, no. 1, pp. 41–45, Jan. 2013.

[16] S. Qaiser and R. Ali, "Text Mining: Use of TF-IDF to Examine the Relevance of Words to Documents," *IJCA*, vol. 181, no. 1, pp. 25–29, Jul. 2018, doi: 10.5120/ijca2018917395.

[17] C. Goutte and E. Gaussier, "A Probabilistic Interpretation of Precision, Recall and F-Score, with Implication for Evaluation," in *Advances in Information Retrieval*, D. E. Losada and J. M. Fernández-Luna, Eds., in Lecture Notes in Computer Science, vol. 3408. Berlin, Heidelberg: Springer Berlin Heidelberg, 2005, pp. 345–359. doi: 10.1007/978-3-540-31865-1_25.

[18] V. Balakrishnan and L.-Y. Ethel, "Stemming and Lemmatization: A Comparison of Retrieval Performances," *LNSE*, vol. 2, no. 3, pp. 262–267, 2014, doi: 10.7763/LNSE.2014.V2.134.